\documentclass[letterpaper]{article} 
\usepackage[preprint]{aaai2027}  
\usepackage[hyphens]{url}  
\usepackage{graphicx} 
\usepackage{natbib}  
\usepackage{caption} 
\usepackage{algorithm}
\usepackage{algorithmic}
\usepackage{array} 
\usepackage[table]{xcolor}
\definecolor{rowblue}{RGB}{220,230,241} 
\usepackage{makecell}
\usepackage{multirow}
\usepackage{pifont}
\newcommand{\name}{\emph{OmniMech}}
 \usepackage{subcaption} 
\usepackage{xcolor}
\usepackage{pifont}
\usepackage{amsmath}
\newcommand{\cmark}{\textcolor{green!60!black}{\ding{51}}}
\newcommand{\xmark}{\textcolor{red!75!black}{\ding{55}}}

\usepackage{newfloat}
\usepackage{listings}
\DeclareCaptionStyle{ruled}{labelfont=normalfont,labelsep=colon,strut=off} 
\floatstyle{ruled}
\newfloat{listing}{tb}{lst}{}
\floatname{listing}{Listing}

\usepackage{booktabs}

\title{OmniMech: All-in-one Multimodal Mechanical Benchmark for 3D Reconstruction }
\author{
    Taiting Lu\textsuperscript{\rm 1}\equalcontrib,
    Runze Liu\textsuperscript{\rm 1}\equalcontrib,
    Ziwei Dong\textsuperscript{\rm 2},
    Sisong Bei\textsuperscript{\rm 2},
    Jingying Zeng\textsuperscript{\rm 2},
    Mingjia Wang\textsuperscript{\rm 3},
    \\
    Zhenghao Li\textsuperscript{\rm 1},
    Kaiyuan Lin\textsuperscript{\rm 1},
    Yi-Shan Wu\textsuperscript{\rm 5},
    Yangshoudu Zheng\textsuperscript{\rm 3},
    Hongxing Pan\textsuperscript{\rm 3},
    Kai Zhang\textsuperscript{\rm 3},
    \\
    Guoliang Shi\textsuperscript{\rm 3},
    Ling Ma\textsuperscript{\rm 3},
    Yifan Yang\textsuperscript{\rm 4},
    Jiaying Lu\textsuperscript{\rm 5},
    Qi He\textsuperscript{\rm 2},
    Sung-Liang Chen\textsuperscript{\rm 3},
    Yi-Chao Chen\textsuperscript{\rm 3},
    Yincheng Jin\textsuperscript{\rm 5},
    Mahanth Gowda\textsuperscript{\rm 1}\corresponding
}

\affiliations{
    \textsuperscript{\rm 1}Pennsylvania State University,
    \textsuperscript{\rm 2}Independent Researcher,
    \textsuperscript{\rm 3}Shanghai Jiao Tong University\\
    \textsuperscript{\rm 4}Microsoft Research,
    \textsuperscript{\rm 5}Binghamton University
}



\begin{document}

\maketitle


\begin{abstract}
Recent vision-language models (VLMs) have demonstrated a remarkable capability to generate executable CAD programs from images by aligning visual geometry with symbolic program representations. 
However, existing approaches primarily focus on coarse, general-purpose 3D object datasets, such as tables and chairs, while overlooking the fine-grained geometry and millimeter-level tolerance requirements of high-fidelity 3D CAD, whose functionality and manufacturability depend on precise dimensions.
To bridge this gap, we introduce \textbf{\name}, the first million-grade benchmark for evaluating VLMs on generating executable CAD programs from industrial mechanical manufacturing.
\textbf{\name} comprises over 251,000 fully dimensioned and toleranced industrial 2D orthographic mechanical projection drawings with semantic annotations, each paired with its corresponding native CAD model, multi-view renderings, 3D geometric representation (mesh, STEP and B-rep), and rich semantic annotations.
The benchmark comprises four tasks: 
\textbf{(1) parametric CAD program synthesis}, benchmarking VLMs alongside CAD-specialized CAD generation models on generating executable parametric CAD programs that construct valid 3D objects from 2D engineering drawings; \textbf{(2) diagram-to-3D reasoning}, evaluating VLMs alongside CAD-specialized models on reconstructing geometry that is visually and structurally consistent with the multi-view evidence in the input drawing; \textbf{(3) annotation-grounded geometric reasoning}, interpreting dimensions, symbols, and feature callouts and enforcing them as geometric constraints in the generated object; and \textbf{(4) spherical visual tool-augmented agentic reasoning}, selecting and invoking visualization, measurement, CAD execution, and verification tools to iteratively accomplish tasks (1)–(3).
Our results reveal that VLMs and even existing state-of-the-art vision exhibit substantial limitations in industrial CAD generation, including unreliable synthesis of executable parametric programs, brittle reconstruction of fine-grained 3D geometry, and poor grounding and enforcement of dimensional annotations and manufacturing tolerances.
We will open-source all benchmark data, evaluation code, and tool interfaces to facilitate future research.

\end{abstract}

\begin{figure*}[!t]
    \centering
    \includegraphics[width=0.90\textwidth]
        {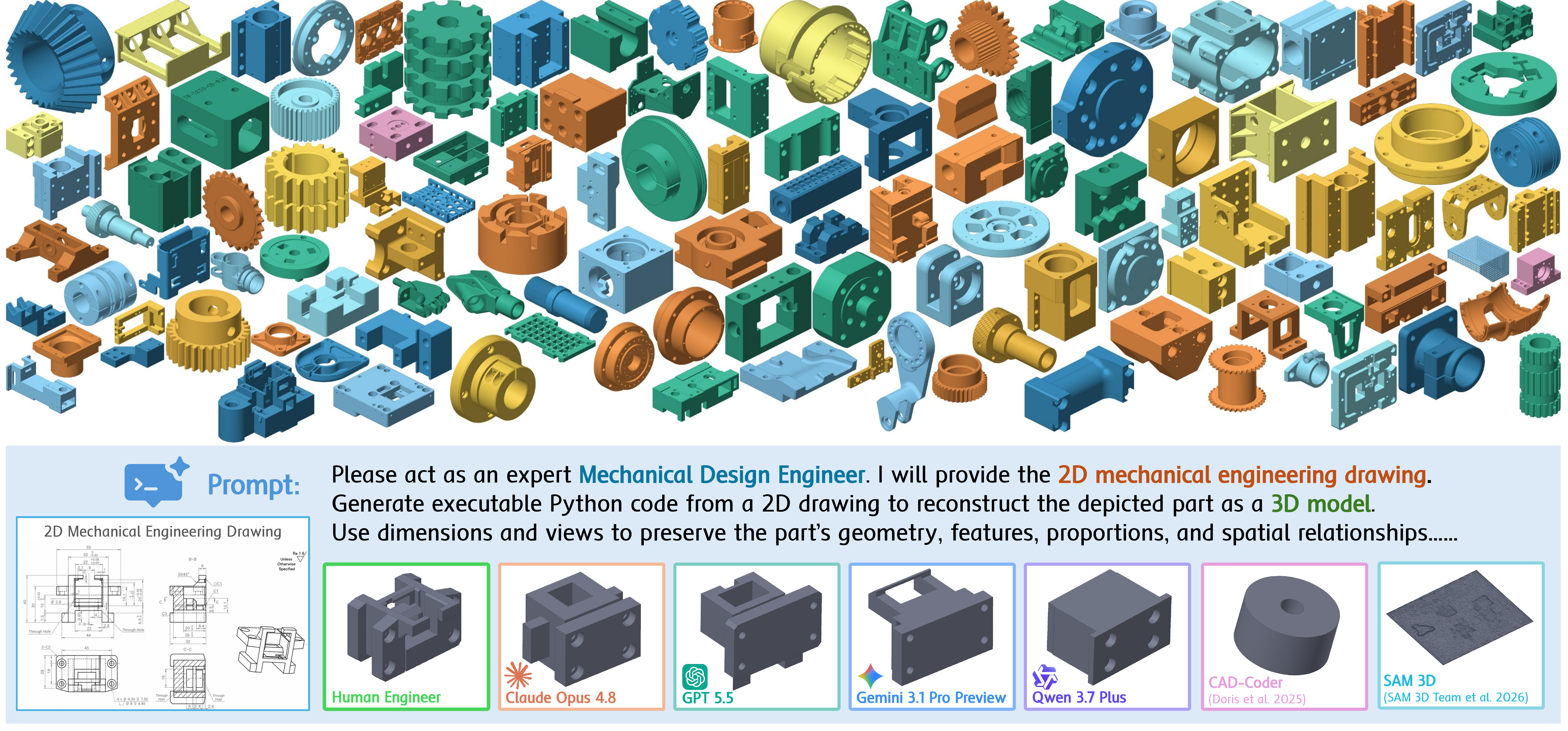}
    \caption{Large language models (LLMs) exhibit limited capability in converting 2D mechanical engineering drawings into accurate 3D models. Frequent reconstruction failures, including omitted features, incorrect feature dimensions, and inconsisten geometric constraints, demonstrate significant shortcomings in spatial reasoning and engineering-aware 3D reconstruction.}
    \label{fig:lmm_test}
\end{figure*}

\section{Introduction}\label{intro}

  

Computer-aided design (CAD) is a foundational technology in modern engineering and industrial design, enabling the creation of precise, editable 3D models of complex objects across products ranging from everyday smartphones to advanced robotic systems. As essential human-readable 2D representations within CAD workflows, orthographic mechanical projection drawings communicate geometry, dimensions, and design intent, in contrast to machine-readable 3D representations such as polygonal meshes and boundary-representation (B-rep) CAD models. However, most real-world mechanical drawings are proprietary and rarely available for public use, while reconstructing their corresponding 3D CAD models still depends heavily on professional engineers to manually interpret the drawings and reproduce the 3D object models. To support downstream CAD workflows, mechanical drawings must therefore be reconstructed as high-fidelity, machine-readable 3D CAD models that faithfully preserve fine-grained geometry, complex but accurate dimensional specifications, topology, and design constraints, enabling reliable analysis, simulation, and manufacturing. Unlike general-purpose 3D reconstruction from natural images \cite{li2025triposg,hunyuan3d2025hunyuan3d,chen2026sam,wu2026direct3d,li2026sldprtnet}, which has been extensively studied with large-scale datasets and typically emphasizes visually plausible geometry and overall shape similarity, reconstructing editable CAD models from mechanical drawings remains comparatively underexplored and requires substantially higher geometric and engineering fidelity.


Recently, several methods, including CME-CAD, CAD-Coder, and CAD2Program \cite{wang20252d,doris2026cad,niu2026cme}, have leveraged vision-language models to reconstruct editable 3D CAD models from 2D mechanical drawings or CAD images by generating executable CAD programs. Despite encouraging progress, the lack of realistic training data remains a major bottleneck. As shown in Table \ref{tab:cad_dataset}, existing public CAD datasets largely lack mechanical drawing annotations, and even the most closely related dataset, CME-CAD \cite{niu2026cme}, provides only three engineering annotations per model. This is far below the tens or even hundreds of annotations typically required to describe the geometric complexity and manufacturing intent of real-world mechanical drawings, limiting current methods’ ability to achieve high-fidelity CAD reconstruction in practical industrial scenarios. Meanwhile, large multimodal models (LMMs) \cite{sun2024dreamcraft3d,liu2023zero,chen2024spatialvlm,niu2026cme,li2026sldprtnet} have demonstrated strong capabilities in text-guided 3D generation, image-conditioned 3D reconstruction, spatial reasoning, and fine-grained geometric understanding, suggesting new opportunities for mechanical drawing-to-3D reconstruction through visual-symbolic alignment, cross-view 3D reasoning, and precise recovery of geometric features and dimensions. Yet it remains unclear whether these capabilities are sufficient for the highly demanding industrial setting, where models must establish correspondences across orthographic views, infer occluded geometry and feature topology, resolve projection ambiguities, and translate design intent into dimensionally consistent parametric CAD operations. This raises a critical open question: \textbf{\textit{Can current LMMs generate executable CAD programs that reconstruct high-fidelity 3D models from industrial mechanical drawings while accurately interpreting engineering semantics and preserving geometric and dimensional consistency across diverse real-world designs?}}

To investigate this question, we conducted a series of preliminary experiments evaluating the mechanical drawing-to-3D reconstruction capabilities of state-of-the-art general-purpose VLMs including Claude Opus 4.8 \cite{anthropic2026claudeopus48}, GPT-5.5 \cite{openai_gpt55}, Gemini 3.1 Pro Preview \cite{google2026gemini31propreview}, and Qwen3.7-Plus \cite{qwen3_7_plus}, alongside CAD-specialized generation models, including CAD-Coder \cite{doris2026cad} and SAM-3D \cite{chen2026sam}.
Their performance was compared with that of experienced mechanical engineers.
Specifically, we evaluated their ability to generate executable CAD programs or 3D models from real-world mechanical drawings paired with engineer-validated 3D models.
As illustrated in Figure \ref{fig:lmm_test}, general-purpose VLMs are unable to reliably interpret engineering annotations or reason across multiple orthographic views, while CAD-specialized models lack the capability to recover complete 3D structures and synthesize correct CAD programs.

\begin{figure*}[t]
    \centering

    \includegraphics[width=0.23\textwidth]
        {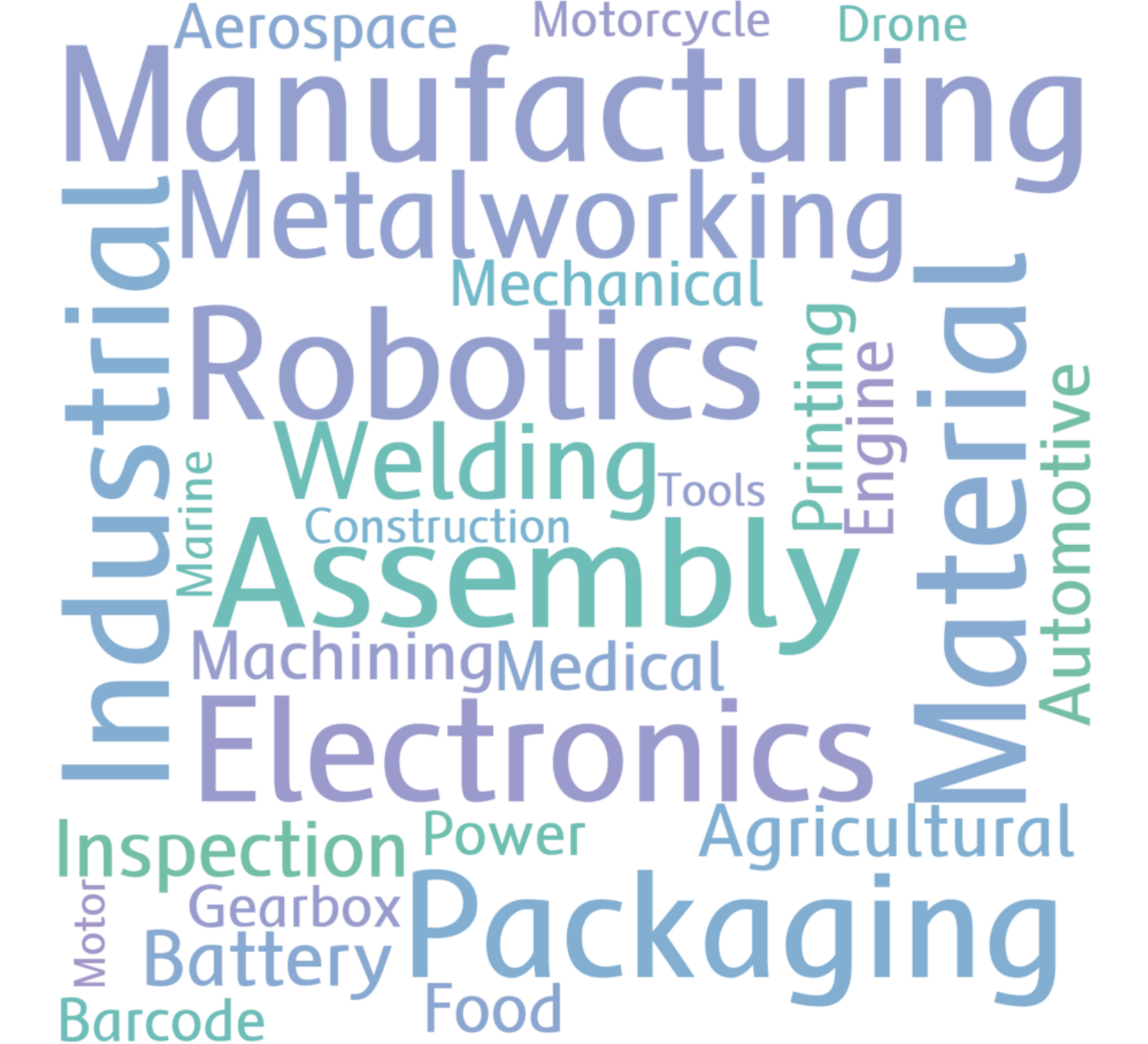}
    \hfill
    \includegraphics[width=0.23\textwidth]
        {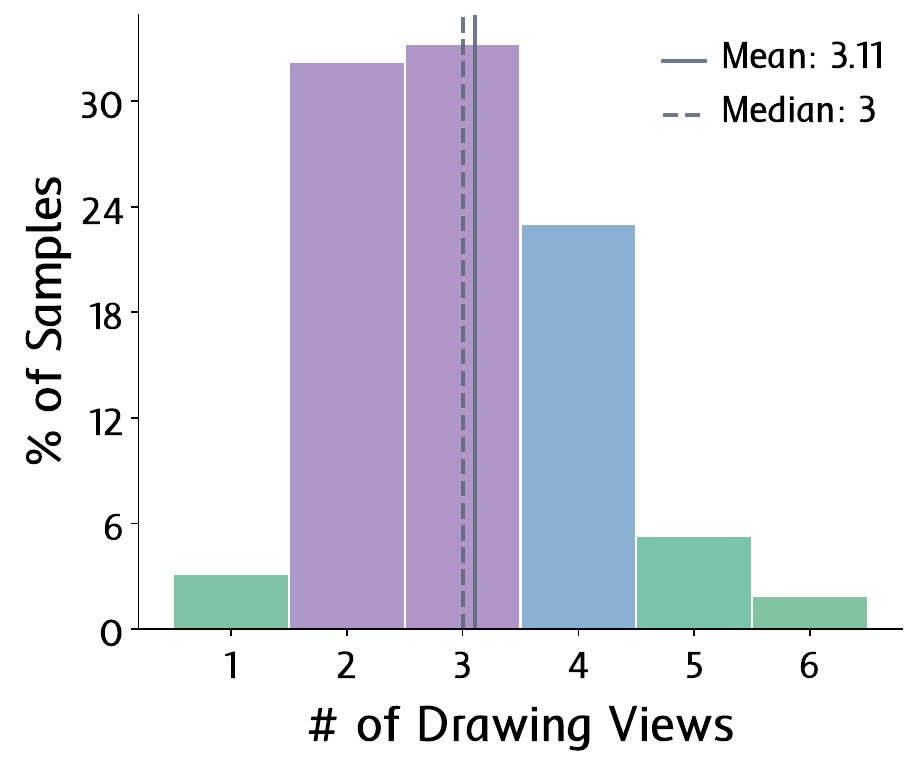}
    \hfill
    \includegraphics[width=0.23\textwidth]
        {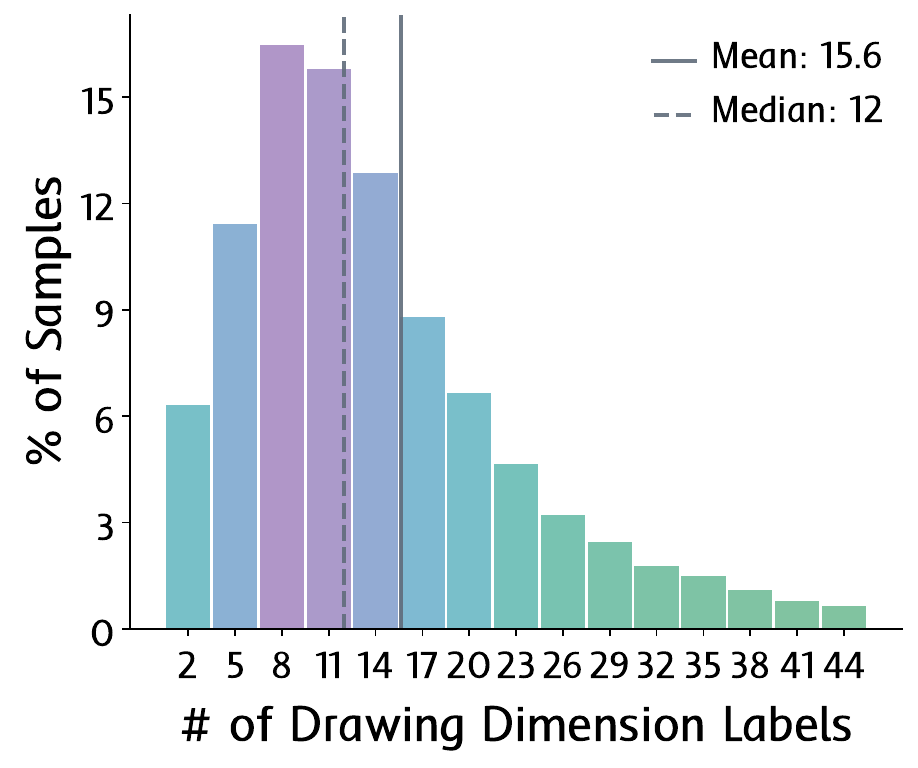}
    \hfill
    \includegraphics[width=0.23\textwidth]
        {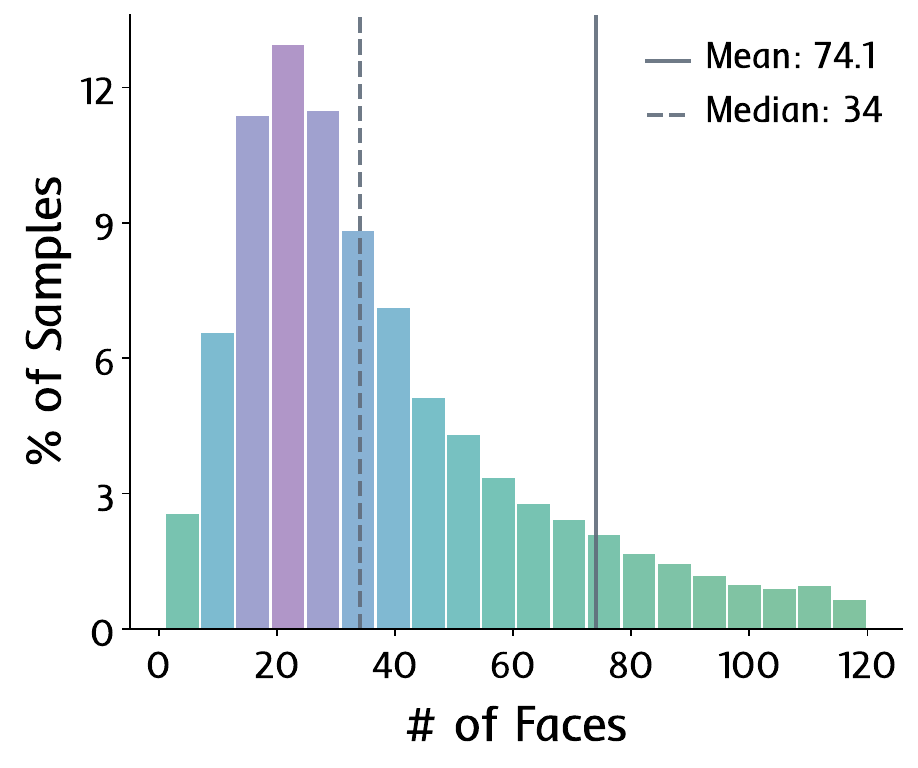}

    \caption{
        Statistical overview of the {\name} benchmark. From left to right: application word cloud, number of views per drawing, number of dimension labels per drawing, and number of faces per model. 
    }
    \label{fig:img2sch_statistic}
\end{figure*}

\begin{figure*}[!ht]
    \centering
    \includegraphics[width=0.85\textwidth]{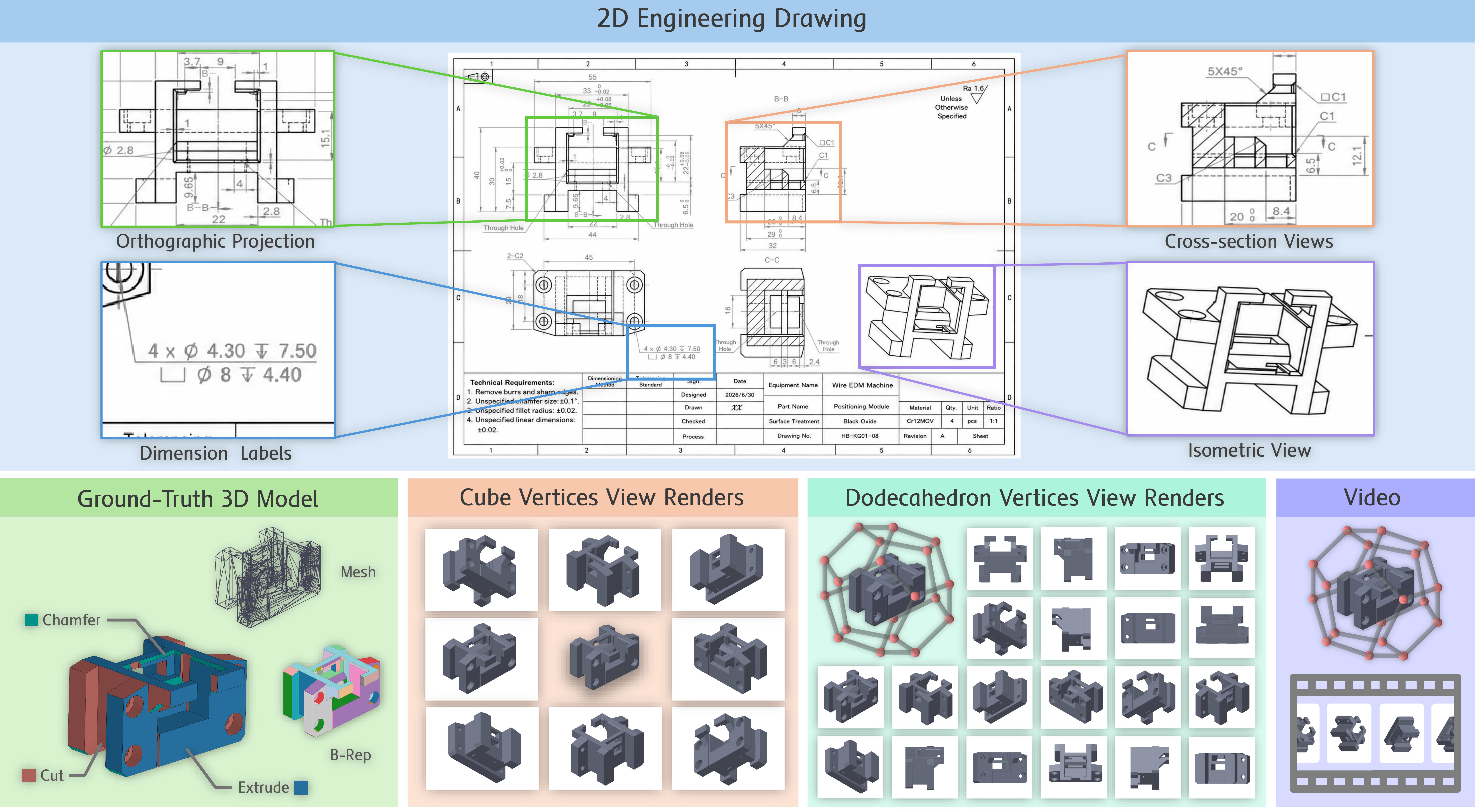}
    \caption{Overview of {\name} benchmark with representative cases.}
    \label{fig:dataset_overview}
\end{figure*}

To bridge this gap, we introduce \textbf{\name}, the first large-scale benchmark for evaluating VLMs on industrial mechanical drawing-to-3D reconstruction through executable CAD program generation. Each instance is centered on a native parametric SolidWorks file (\texttt{.sldprt}), preserving CAD-interface editability and supporting conversion to multiple formats, with accompanying B-Rep (\texttt{.step}), mesh (\texttt{.obj}), 20-views spherical renderings, and a feature graph.
As shown in Figure \ref{fig:dataset_overview}, \textbf{\name} contains 251K real-world mechanical drawings paired with corresponding multi-format 3D representations, spanning a broad range of industrial application domains, including robotics, automotive systems, aerospace equipment, and medical components.
The benchmark comprises 251K mechanical drawing paired 3D models.
We first propose a systematic evaluation protocol for mechanical drawing-to-3D reconstruction via executable CAD program generation, assessing performance from three complementary perspectives: parametric CAD program synthesis, diagram-to-3D geometric consistency, and annotation-grounded constraint reasoning.
Our benchmark evaluates four core capabilities:
\textbf{(i) parametric CAD program synthesis} for generating executable programs that construct valid 3D objects from 2D mechanical drawings;
\textbf{(ii) diagram-to-3D reasoning} for evaluating the geometric, structural, and topological fidelity of the 3D models produced by executing generated CAD programs against the corresponding ground-truth 3D models;
\textbf{(iii) annotation-grounded geometric reasoning} for interpreting dimensions, symbols, and feature callouts and enforcing them as constraints in the reconstructed model;
and \textbf{(iv) tool-augmented agentic reconstruction} for iteratively selecting and invoking visualization, measurement, CAD execution, and verification tools to refine the generated 3D CAD model.

Our main contributions are as follows:
\textbf{(i)} We introduce \textbf{\name}, a benchmark of 251K real-world mechanical drawings paired with editable parametric CAD models and multi-format 3D ground truth, together with systematic protocols for evaluating CAD program generation and reconstruction fidelity.
\textbf{(ii)} We systematically evaluate state-of-the-art general-purpose LMMs and CAD-specialized generation models under multiple settings, including zero-shot CAD program synthesis and tool-augmented agentic reconstruction with iterative visualization, measurement, CAD execution, and verification.
\textbf{(iii)} We provide a detailed analysis of current models across parametric CAD program synthesis, diagram-to-3D reasoning, annotation-grounded geometric reasoning, and tool-augmented refinement, revealing their key limitations in producing accurate, editable, and dimensionally consistent 3D CAD models.

\begin{table}[h!]
\centering
\begingroup
\fontsize{6}{6.5}\selectfont
\setlength{\tabcolsep}{3pt}
\renewcommand{\arraystretch}{1}
\begin{tabular}{
m{2.7cm}
>{\centering\arraybackslash}m{0.4cm}
>{\centering\arraybackslash}m{0.8cm}
>{\centering\arraybackslash}m{0.5cm}
>{\centering\arraybackslash}m{0.6cm}
>{\centering\arraybackslash}m{0.6cm}
>{\centering\arraybackslash}m{0.4cm}
>{\centering\arraybackslash}m{0.75cm}
}
\toprule
\multirow{2}{*}{\textbf{Dataset}} &
\multirow{2}{*}{\makecell{\textbf{Mech.}\\\textbf{Drw.}}} &
\multirow{2}{*}{\makecell{\textbf{Avg. Eng.}\\\textbf{Anno.}}} &
\multirow{2}{*}{\textbf{\#View}} &
\multirow{2}{*}{\textbf{\#Model}} &
\multicolumn{3}{c}{\textbf{3D Object}} \\
\cmidrule(lr){6-8}
& & & & &
\textbf{B-Rep} &
\textbf{Mesh} &
\textbf{Param.}
\\
\midrule
\shortstack[l]{\textbf{PrincetonSB} {\fontsize{5.5}{6}\selectfont \cite{shilane2004princeton}}}  & \xmark & \xmark & \xmark & 6.67K    & \xmark & \cmark & \xmark  \\ 
\specialrule{0.01em}{0.1ex}{0.5ex}
\shortstack[l]{\textbf{AAD} {\fontsize{5.5}{6}\selectfont\cite{bespalov2005benchmarking}}}      & \xmark & \xmark & \xmark & 180      & \cmark & \cmark & \xmark  \\ 
\specialrule{0.01em}{0.1ex}{0.5ex}
\shortstack[l]{\textbf{ESB} {\fontsize{5.5}{6}\selectfont\cite{jayanti2006developing}}}        & \xmark & \xmark & \xmark & 867      & \cmark & \cmark & \xmark  \\ 
\specialrule{0.01em}{0.1ex}{0.5ex}
\shortstack[l]{\textbf{ModelNet} {\fontsize{5.5}{6}\selectfont\cite{wu20153d}}}                 & \xmark & \xmark & \xmark & 151.13K  & \xmark & \cmark & \xmark  \\ 
\specialrule{0.01em}{0.1ex}{0.5ex}
\shortstack[l]{\textbf{ShapeNet} {\fontsize{5.5}{6}\selectfont\cite{chang2015shapenet}}}        & \xmark & \xmark & \xmark & 3000K    & \xmark & \cmark & \xmark  \\ 
\specialrule{0.01em}{0.1ex}{0.5ex}
\shortstack[l]{\textbf{PartNet} {\fontsize{5.5}{6}\selectfont\cite{mo2019partnet}}}             & \xmark & \xmark & \xmark & 26.67K   & \xmark & \cmark & \xmark  \\ 
\specialrule{0.01em}{0.1ex}{0.5ex}
\shortstack[l]{\textbf{ABC} {\fontsize{5.5}{6}\selectfont\cite{koch2019abc}}}                   & \xmark & \xmark & \xmark & 1000K    & \cmark & \cmark & \xmark  \\ 
\specialrule{0.01em}{0.1ex}{0.5ex}
\shortstack[l]{\textbf{MCB} {\fontsize{5.5}{6}\selectfont\cite{kim2020large}}}                  & \xmark & \xmark & \xmark & 58.7K    & \xmark & \cmark & \xmark  \\ 
\specialrule{0.01em}{0.1ex}{0.5ex}
\shortstack[l]{\textbf{Fusion 360} {\fontsize{5.5}{6}\selectfont\cite{willis2021fusion}}}       & \xmark & \xmark & \xmark & 8.62K    & \cmark & \cmark & DSL     \\ 
\specialrule{0.01em}{0.1ex}{0.5ex}
\shortstack[l]{\textbf{DeepCAD*} {\fontsize{5.5}{6}\selectfont (Wu et al.~\citeyear{wu2021deepcad})}} & \xmark & \xmark & \xmark & 178.24K  & \xmark & \xmark & DSL \\ 
\specialrule{0.01em}{0.1ex}{0.5ex}
\shortstack[l]{\textbf{CAD-Coder*} {\fontsize{5.5}{6}\selectfont\cite{doris2026cad}}}           & \xmark & \xmark & 5      & 163.67K  & \xmark & \xmark & CadQuery \\ 
\specialrule{0.01em}{0.1ex}{0.5ex}
\shortstack[l]{\textbf{CME-CAD} {\fontsize{5.5}{6}\selectfont\cite{niu2026cme}}}                & \cmark & 2      & 3      & 17.3K    & \xmark & \xmark & CadQuery \\  
\specialrule{0.01em}{0.1ex}{0.5ex}
\rowcolor{rowblue}\shortstack[l]{\textbf{OmniMech} \textbf{(Our)}}
& \textbf{\cmark}
& \textbf{16}
& \textbf{3}
& \textbf{251K}
& \textbf{\cmark}
& \textbf{\cmark}
& \textbf{Sldwrks.}\\
\bottomrule
\end{tabular}
\caption{
Comparison of publicly available CAD datasets. Mech. Drw. refers to 2D mechanical engineering drawing; Avg. Eng. Anno. refers to average drawing dimension annotation; Sldwrks. refers to SolidWorks data; Param. refers to type of parametric models;
DSL refers to domain-specific language representations for parametric CAD models. 
An asterisk (*) indicates datasets derived from the ABC dataset.
}
\label{tab:cad_dataset}

\endgroup
\end{table}

\textbf{3D Reconstruction from Image.}
The problem of 3D reconstruction from images is long-standing in computer vision, computer graphics, and CAD. Among the earliest works on image-based 3D reconstruction, \cite{roberts1963machine} and \cite{horn1970shape} formulated single-image 3D reconstruction as a principled computational problem, with Roberts inferring three-dimensional polyhedral scene structure from two-dimensional visual evidence and Horn recovering smooth surface geometry from image shading. Subsequent studies \cite{choy20163d, wang2018pixel2mesh, xie2019pix2vox, saito2019pifu,chen2026sam} explored increasingly expressive 3D representations, enabling direct reconstruction from a single image to voxels, meshes, and implicit surfaces. More recently, several works \cite{wang2026shapellm,fan2026vlm,hu2026g} integrate semantic reasoning from VLMs with geometric modeling to jointly understand visual content and infer three-dimensional structure, thereby improving generalization and spatial reasoning.

\section{Related Work}\label{related_work}

\textbf{Benchmark for Image-to-3D reconstruction.}
Over the past decade, image-to-3D reconstruction has been an active area of research, resulting in the establishment of a variety of public datasets and standardized evaluation benchmarks.
As shown in Table \ref{tab:cad_dataset}, early datasets such as PrincetonSB \cite{shilane2004princeton}, ModelNet \cite{wu20153d}, and ShapeNet \cite{chang2015shapenet} established large-scale mesh repositories for 3D shape understanding, primarily focusing on general-purpose objects such as chairs and tables.
ABC \cite{koch2019abc} extended this direction to one million mechanical CAD models with B-Rep and mesh representations. Building on ABC, DeepCAD \cite{wu2021deepcad} derived editable parametric construction sequences, while CAD-Coder \cite{doris2026cad} further paired 163.67K derived CadQuery models with five rendered views. More recently, CME-CAD \cite{niu2026cme} pairs 17.3K models with synthetic mechanical drawings, three views, and only 2--3 engineering annotations per drawing.
Existing datasets mainly formulate image-to-3D reconstruction as multi-view shape recovery with parametric programs, but their rendered inputs lack semantic and engineering supervision, such as dimensions, tolerances, and feature-level annotations. Consequently, current methods prioritize visual and geometric similarity over dimensional accuracy, topological correctness, and engineering validity. This remains insufficient for mechanical engineering, which requires exact geometry, manufacturable structures, and hundreds of annotations per model.

\section{Benchmark Construction}\label{benchmark}

\textbf{Task Formulation.}
To provide a comprehensive evaluation framework for image-to-3D reconstruction, our task formulation covers four key capabilities of multimodal geometric reasoning:
\textbf{(i)} \textbf{Parametric CAD Program Reasoning}, which requires translating 2D mechanical drawings into executable CAD programs with valid operation sequences, geometric dependencies, and editable parametric structures;
\textbf{(ii)} \textbf{3D Spatial Reasoning}, which requires inferring complete three-dimensional geometry, topology, and occluded structures from multi-view 2D observations, evaluated by the fidelity of the reconstructed model to the ground-truth shape;
\textbf{(iii)} \textbf{Symbolic Geometric Reasoning}, which requires interpreting dimensions, symbols, and feature callouts and enforcing them as explicit geometric constraints in the reconstructed CAD model;
\textbf{(iv)} \textbf{Tool-Augmented Agentic 3D Reconstruction}, which formulates reconstruction as an iterative decision-making process, where VLMs invoke visualization, measurement, CAD execution, and verification tools to progressively refine the model until geometric, structural, and annotation constraints are satisfied.

\textbf{Annotation Curation.}
To construct {\name}, we collect real-world editable 3D CAD assembly models from open-access online repositories, including GrabCAD, 3D ContentCentral, TraceParts, and GitHub~\cite{grabcad,contentcentral,traceparts,github}. 
The collected projects are created using SolidWorks~\cite{solidworks}, a widely used commercial CAD software.
Each 3D model consists of a native SolidWorks part file (\texttt{.SLDPRT}) paired with its corresponding orthographic mechanical drawing file (\texttt{.SLDDRW}), which captures dimensional annotations, geometric specifications, and design intent.
To support diverse geometric learning paradigms, we use the SolidWorks API~\cite{solidworks_api} to export each part into multiple representations, including a STEP B-Rep (\texttt{.STEP}), an OBJ mesh (\texttt{.OBJ}), 28 rendered views, including 20 from dodecahedron vertices and 8 from cube vertices, and a walkthrough video for each 3D model.

\textbf{Statistics of {\name} Benchmark.} 
As illustrated in Figure~\ref{fig:img2sch_statistic} and Figure~\ref{fig:dataset_overview}, {\name} contains 251K real-world 3D mechanical designs paired with corresponding engineering drawings. 
The dataset spans a wide range of application domains, including robotics, agricultural systems, drones, and other related fields.
Our dataset includes three levels of representation:
\textbf{(i)} \textbf{Mechanical Drawing-Coupled 3D Model Design}, comprising 251K 3D models with a total of $37,071$ solids, $2,048,082$ faces, $3,476,060$ vertices, and volumes ranging from $1,237$ to $2,145,030$~$\mathrm{mm}^3$. Each model is paired with a corresponding orthographic mechanical drawing containing, on average, $15.62$ dimensional annotations.
\textbf{(ii)} \textbf{Geometric Representations}, including the native editable parametric (\texttt{.SLDPRT}), STEP B-Rep (\texttt{.STEP}), and OBJ mesh (\texttt{.OBJ}).
\textbf{(iii)} \textbf{Visual Representations}, with 28 rendered multi-view images and a walkthrough video for each 3D model. Detailed depiction of dataset composition is arranged in Appendix A due to space constraint.


\section{Evaluation Metrics}

\textbf{Evaluation Protocol.}

\textbf{Evaluation Criteria.}
We define evaluation metrics across four complementary dimensions to comprehensively assess mechanical drawing-to-3D model quality; full details of each metric are provided in Appendix B due to space constraints.
\textbf{(i)} \textbf{CAD Program Legality}, which evaluates the syntactic validity and executability of the generated CAD program, together with its program length and the diversity of the CAD operations employed;
\textbf{(ii)} \textbf{3D Model Quality}, which assesses the geometric fidelity of the 3D model reconstructed from the executable CAD program, measured by: (a) \textit{Surface Similarity}, quantified using Chamfer Distance (CD), surface intersection-over-union (SIoU), and Surface F-score; (b) \textit{Volumetric Similarity}, measured using volumetric intersection-over-union (VIoU); (c) \textit{Surface Area Accuracy}, evaluated through absolute and relative surface-area errors; and (d) \textit{Volume Accuracy}, evaluated through absolute and relative volume errors;
\textbf{(iii)} \textbf{Agentic Evaluation}, which assesses the ability of LMMs to iteratively refine CAD programs for 3D reconstruction through visual tool use, measured by: (a) \textit{Inference Efficiency}, the total time required to complete the reconstruction process; (b) \textit{Token Efficiency}, the number of output tokens generated during iterative refinement; and (c) \textit{Tool-Use Efficiency}, the average number of tool-use steps required to produce a valid 3D model.

\section{Experiment and Findings}

\begin{figure}[t]
    \centering
    \includegraphics[width=0.9\columnwidth]{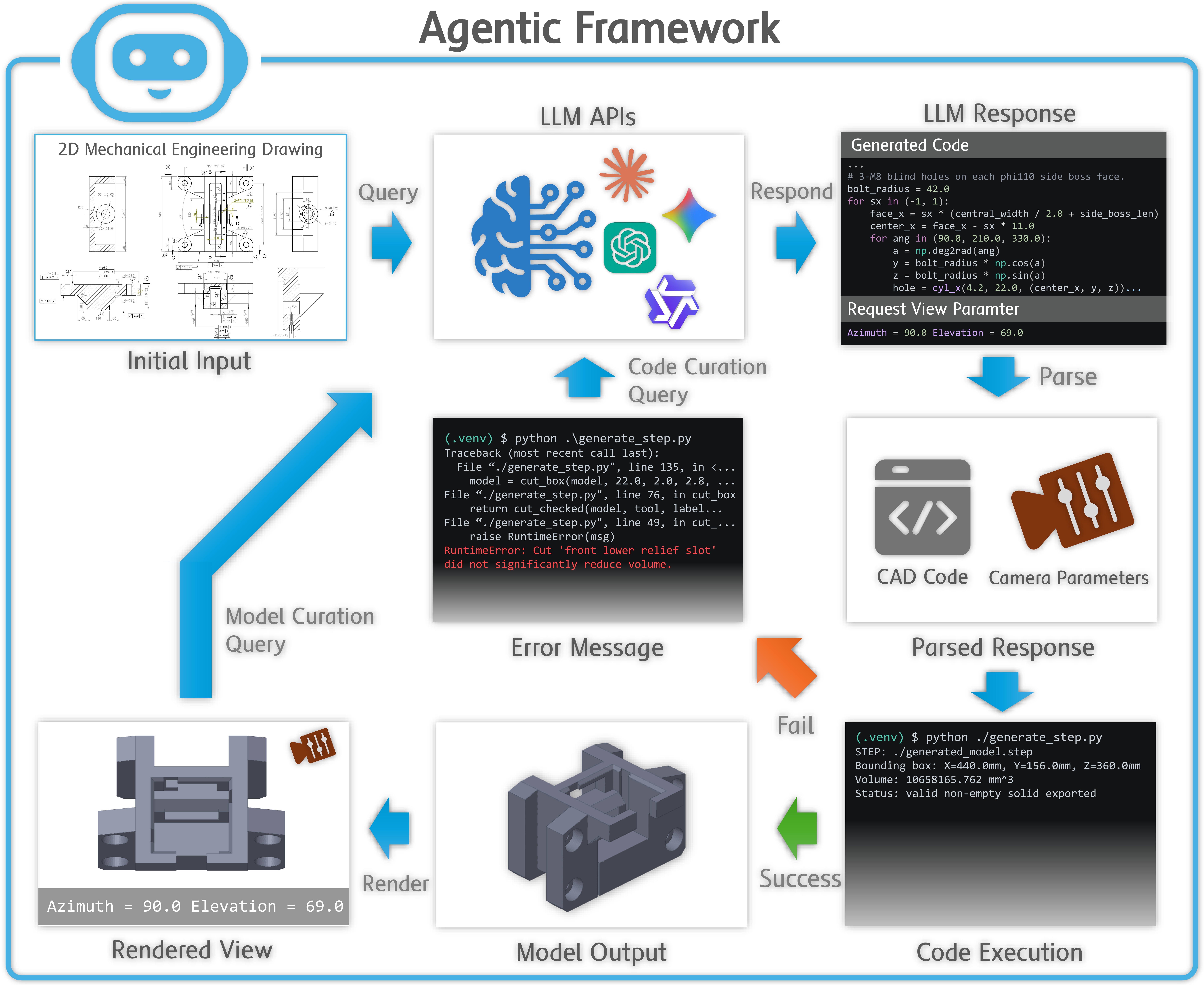}
    \caption{Overview of an agentic framework for evaluating LMM-based tool usage in mechanical model reconstruction.}
    \label{fig:react_framework}
\end{figure}

\subsection{Experimental Setups}\label{sec:exp}

\textbf{Study Setup.} The tested LMMs in this section include GPT-5.5~\cite{openai_gpt55}, GPT-5.4 mini~\cite{openai_gpt5mini}, Gemini 3.1 Pro Preview~\cite{google_gemini3_pro}, Gemini 3.5 Flash~\cite{googledeepmind2026gemini35flash}, Claude Opus 4.8~\cite{anthropic2026claudeopus48}, LLaMA 4 Maverick~\cite{meta_llama4}, Mistral 3 14B 25.12 ~\cite{mistralai2025ministral314b}, Qwen 3.5-9B~\cite{qwen35_9b}, and Qwen 3.7-Plus~\cite{qwen3_7_plus}. 
All settings use a single mechanical drawing as base input. We further investigate the impact of additional visual information by incorporating 8 cubic-view images, or a dodecahedron walkthrough video. For VLMs that do not support video input, we use 20 dodecahedron-view images as an alternative visual representation of the walkthrough video. We also evaluate the effect of few-shot examples; and \textbf{(ii)} \textbf{Agentic Evaluation}, where LMMs are equipped with a 3D rendering interface to iteratively refine the reconstructed model. We conduct two ablations: (a) \textit{Adaptive View Selection}, where the LMM dynamically controls the camera angle and viewing distance to inspect fine-grained geometric details; and (b) \textit{Fixed Multi-View Observation}, where the LMM is provided with eight predefined cubic-view images upon request.

\textbf{Implementation of Agentic Framework.} 
We design a multi-agent, multi-round agentic evaluation framework in which LMMs are provided with a mechanical drawing and an initial CAD program, together with two visual inspection tools: adaptive view selection, which allows the model to control the camera angle and viewing distance, and fixed multi-view observation, which provides eight predefined cubic-view renderings upon request. At each step, the LMM takes one of two actions based on the local execution outcome: if the Python script produces an execution error, the LMM revises the code according to the error message; otherwise, it inspects the rendered 3D model and iteratively refines the modeling operations, dimensions, and feature parameters. This process continues until a valid reconstruction is obtained or the maximum number of steps is reached.

\textbf{Specialized CAD Generation Baselines.}
To comprehensively evaluate the difficulty and quality of our benchmark, we consider both direct image-to-3D and program-based image-to-CAD reconstruction. For direct image-to-3D reconstruction, we evaluate SAM-3D~\cite{chen2026sam}, which predicts object geometry, texture, and layout; TRELLIS~\cite{xiang2024structured}, which generates multiple 3D representations through structured latents; InstantMesh~\cite{xu2024instantmesh}, which combines multi-view diffusion with sparse-view mesh reconstruction; and Hunyuan3D~\cite{hunyuan3d22025tencent}, which separately generates geometry and texture. For program-based image-to-CAD reconstruction, we evaluate CAD-Coder~\cite{doris2026cad}, which generates executable CadQuery programs for editable parametric models.

\textbf{Compressed Agentic Context}
In the agentic setting, the model generates CAD construction scripts at every iteration. To preserve iterative context without resending all previous scripts in full, we design a compressed code-history representation. The most recent script is provided complete, while earlier scripts are represented by diffs from the latest script to the immediately preceding version. This preserves the model's access to prior design decisions, while reducing duplicated source-code tokens across iterative refinement turns.

Additional experiment setup and computing resource information are provided in Appendix C.

\begin{table*}[h]
\centering

\begingroup
\small
\fontsize{9}{8}\selectfont
\setlength{\tabcolsep}{3pt}
\renewcommand{\arraystretch}{1}
\begin{tabular}{
m{1.5cm}                  
>{\centering\arraybackslash}m{2.4cm} 
>{\centering\arraybackslash}m{0.6cm} 
>{\centering\arraybackslash}m{0.6cm} 
>{\centering\arraybackslash}m{0.6cm} 
>{\centering\arraybackslash}m{0.85cm} 
>{\centering\arraybackslash}m{0.9cm} 
>{\centering\arraybackslash}m{0.9cm} 
>{\centering\arraybackslash}m{1.1cm} 
>{\centering\arraybackslash}m{0.9cm} 
>{\centering\arraybackslash}m{0.9cm} 
> {\centering\arraybackslash}m{0.6cm} 
>{\centering\arraybackslash}m{0.8cm} 
>{\centering\arraybackslash}m{0.85cm} 
>{\centering\arraybackslash}m{0.5cm} 
}
\toprule
& & \multicolumn{4}{c}{\textbf{Geometric Fidelity}}
& \multicolumn{4}{c}{\textbf{Dimensional Fidelity}}
& \multicolumn{5}{c}{\textbf{Inference Statistics}} \\
\cmidrule(lr){3-6}
\cmidrule(lr){7-10}
\cmidrule(lr){11-15}
\multirow{2}{*}{\textbf{Model}} &
\multirow{2}{*}{\textbf{Input}} &
\multirow{2}{*}{\makecell{\textbf{CD}}} &
\multirow{2}{*}{\makecell{\textbf{SIoU}}} &
\multirow{2}{*}{\makecell{\textbf{VIoU}}} &
\multirow{2}{*}{\makecell{\textbf{Surf.}\\\textbf{F-Scr}}} &
\multirow{2}{*}{\makecell{\textbf{Surf.}\\\textbf{Err.}}} &
\multirow{2}{*}{\makecell{\textbf{Surf.}\\\textbf{Err.}}} &
\multirow{2}{*}{\makecell{\textbf{Vol.}\\\textbf{Err.}}} &
\multirow{2}{*}{\makecell{\textbf{Vol.}\\\textbf{Err.}}} &
\multirow{2}{*}{\makecell{\textbf{Infer.}\\\textbf{Time}}} &
\multirow{2}{*}{\makecell{\textbf{Pass}\\\textbf{Rate}}} &
\multirow{2}{*}{\makecell{\textbf{Input}\\\textbf{Tks.}}} &
\multirow{2}{*}{\makecell{\textbf{Output}\\\textbf{Tks.}}} &
\multirow{2}{*}{\makecell{\textbf{CAD}\\\textbf{Ops.}}}\\

\\
 & & $\downarrow$ & $\uparrow$ & $\uparrow$ & $\uparrow$ &
(mm$^2$)$\downarrow$ & (\%)$\downarrow$ &
(mm$^3$)$\downarrow$ & (\%)$\downarrow$ &
(s)$\downarrow$ & (\%)$\uparrow$ & --- & --- & --- \\
\midrule
\multicolumn{15}{c}{\textbf{Commercial LLMs}} \\
\midrule

\multirow[l]{3}{=}{\makecell[l]{GPT\\5.5}}& 2D Draw. & 3.34 & 0.59 & 0.80 & 0.74 & 39450 & 12.5 & 480810 & 41.6 & 309.7 & 92.6 & 2010 & 14864 & 9.1 \\
& Draw.+8 Render & 2.86 & 0.60 & 0.81 & 0.75 & 27572 & 9.6 & 513727 & 25.8 & 288.1 & 91.5 & 10001 & 15687 & 9.2 \\
 & Draw.+20 Render & \textbf{2.80} & 0.62 & 0.82 & 0.76 & 29395 & \underline{9.2} & 664111 & 31.5 & 307.5 & 90.8 & 21745 & 15186 & 9.3 \\
\midrule

 \multirow[l]{3}{=}{\makecell[l]{GPT\\5.4 Mini}}& 2D Draw. & 7.25 & 0.38 & 0.60 & 0.55 & 96050 & 26.1 & 2489940 & 126.7 & 7.5 & 60.0 & 2058 & 1192 & 16.3 \\
& Draw.+8 Render & 6.19 & 0.39 & 0.62 & 0.56 & 54420 & 24.7 & 912537 & 97.0 & 8.4 & 53.0 & 9963 & 1134 & 17.0 \\
 & Draw.+20 Render & 6.56 & 0.40 & 0.63 & 0.57 & 118089 & 27.3 & 1701826 & 145.8 & 10.0 & 48.9 & 21607 & 1122 & 17.1 \\
\midrule

 \multirow[l]{3}{=}{\makecell[l]{Claude\\Opus 4.8}}& 2D Draw. & 4.18 & 0.52 & 0.73 & 0.68 & 49150 & 12.0 & 488890 & 36.9 & 35.8 & 92.7 & 5045 & 2774 & 11.0 \\
& Draw.+8 Render & 3.88 & 0.52 & 0.75 & 0.69 & 45954 & 11.6 & 486893 & 36.2 & 37.3 & \underline{94.2} & 18370 & 2935 & 11.7 \\
 & Draw.+20 Render & 3.72 & 0.52 & 0.75 & 0.69 & 30769 & 10.6 & 367522 & 31.8 & 38.8 & \textbf{94.6} & 36369 & 2925 & 11.7 \\
\midrule

 \multirow[l]{3}{=}{\makecell[l]{Gemini\\3.1 Pro}}& 2D Drawing & 3.46 & 0.64 & 0.83 & 0.78 & 41110 & 11.1 & 1132270 & 23.7 & 171.7 & 81.1 & 2172 & 22037 & 12.9 \\
& Draw.+8 Render & 3.59 & 0.62 & 0.82 & 0.76 & 46574 & \textbf{9.2} & 926373 & \underline{19.0} & 175.9 & 82.3 & 11342 & 22628 & 13.3 \\
 & Draw.+Video & 3.12 & \textbf{0.66} & \underline{0.84} & \textbf{0.80} & 34493 & 12.4 & 1324694 & 22.0 & 182.0 & 80.1 & 2855 & 21917 & 13.1 \\
\midrule

 \multirow[l]{3}{=}{\makecell[l]{Gemini\\3.5 Flash}}& 2D Drawing & 3.49 & 0.63 & 0.84 & 0.77 & 41320 & 9.7 & 276720 & 18.1 & 57.2 & 70.6 & 2192 & 13541 & 13.5 \\
& Draw.+8 Render & 3.01 & \underline{0.66} & \textbf{0.84} & \underline{0.79} & 41321 & 10.8 & 993169 & \textbf{17.7} & 92.1 & 68.0 & 11458 & 14465 & 13.6 \\
 & Draw.+Video & 3.95 & 0.64 & 0.84 & 0.78 & 37989 & 16.8 & 1071291 & 20.9 & 62.9 & 69.3 & 2885 & 14771 & 13.8 \\
\midrule

 \multirow[l]{3}{=}{\makecell[l]{Qwen\\3.7 Plus}}& 2D Drawing & 6.41 & 0.45 & 0.68 & 0.61 & 79300 & 19.1 & 500940 & 56.0 & 226.6 & 66.8 & 3306 & 12203 & 14.5 \\
& Draw.+8 Render & 6.19 & 0.45 & 0.68 & 0.61 & 51614 & 16.8 & 1546826 & 55.0 & 293.1 & 71.2 & 13649 & 15658 & 14.0 \\
 & Draw.+20 Render & 6.97 & 0.46 & 0.71 & 0.63 & 56536 & 14.0 & 1535596 & 79.4 & 322.4 & 70.9 & 28973 & 17223 & 13.6 \\
\midrule

\multicolumn{15}{c}{\textbf{Open-Source LLMs}} \\
\midrule

 \multirow[l]{3}{=}{\makecell[l]{Qwen\\3.5 9B}}& 2D Drawing & 22.2 & 0.40 & 0.60 & 0.55 & 139390 & 237.4 & 3208810 & 448.1 & 403.6 & 1.3 & 3354 & 19773 & 14.7 \\
& Draw.+8 Render & 7.35 & 0.64 & 0.80 & 0.78 & 60451 & 17.6 & 7943561 & 153.1 & 329.1 & 1.0 & 14037 & 19533 & 16.2 \\
 & Draw.+20 Render & 5.79 & 0.32 & 0.65 & 0.48 & 84406 & 37.7 & 2002554 & 277.4 & 350.5 & 1.2 & 29608 & 13342 & 14.2 \\
\midrule

 \multirow[l]{3}{=}{\makecell[l]{Llama4\\Maverick}}& 2D Drawing & 10.3 & 0.33 & 0.52 & 0.48 & 90250 & 39.6 & 3927090 & 283.4 & 9.6 & 61.0 & 5320 & 757 & 8.6 \\
& Draw.+8 Render & 6.41 & 0.36 & 0.57 & 0.52 & 50580 & 27.7 & 1645873 & 186.8 & 110.7 & 45.9 & 14547 & 10427 & 7.9 \\
 & Draw.+20 Render & 7.68 & 0.35 & 0.57 & 0.51 & 67597 & 30.4 & 3107624 & 184.4 & 18.9 & 55.4 & 19067 & 751 & 8.2 \\
\midrule

 \multirow[l]{3}{=}{\makecell[l]{Ministral3\\14B 2512}}& 2D Drawing & 13.9 & 0.29 & 0.52 & 0.44 & 49910 & 23.2 & 391620 & 102.5 & 17.2 & 0.4 & 3351 & 1977 & 14.7 \\
& Draw.+8 Render & 5.87 & 0.31 & 0.46 & 0.45 & 19298 & 77.2 & 123961 & 191.0 & 25.7 & 0.6 & 15169 & 2028 & 15.1 \\
 & Draw.+20 Render & 7.95 & 0.35 & 0.48 & 0.52 & 45539 & 104.7 & 818068 & 287.0 & 28.2 & 0.4 & 15166 & 1923 & 14.9 \\
\midrule

\multicolumn{15}{c}{\textbf{Open-Source 3D and CAD Reconstruction Models}} \\
\midrule

\rlap{HY3D-2 \cite{hunyuan3d2025hunyuan3d}} &   & 18.4 & 0.10 & 0.07 & 0.15 & 144070 & 76.2 & 1233644 & 109.4 & 56.3 & 100 & -- & -- & -- \\
\rlap{InstantMesh \cite{xu2024instantmesh}} &   & 34.1 & 0.04 & 0.05 & 0.06 & 326796 & 200.5 & 42453471 & 5515.5 & 110.9 & 100 & -- & -- & -- \\
\rlap{SAM 3D \cite{chen2026sam}} &   & 15.0 & 0.18 & 0.15 & 0.26 & 206808 & 130.1 & 2426019 & 218.3 & 34.5 & 100 & -- & -- & -- \\
\rlap{TRELLIS \cite{xiang2024structured}} &   & 20.4 & 0.10 & 0.10 & 0.15 & 201123 & 115.4 & 10019880 & 317.9 & 10.2 & 100 & -- & -- & -- \\


\rlap{Cad Coder \cite{doris2026cad}} &   & 16.3 & 0.14 & 0.20 & 0.21 & 107364 & 171.8 & 2275856 & 1395.1 & 126.8 & 28.9 & -- & -- & -- \\
\midrule
\bottomrule
\end{tabular}
\caption{
Zero-shot Experiment Results (Sample N=5000).
}
\label{tab:oneshot_result}

\endgroup
\end{table*}

\begin{table*}[t]
\centering

\begingroup
\fontsize{9}{8.5}\selectfont
\setlength{\tabcolsep}{3pt}
\renewcommand{\arraystretch}{1}
\begin{tabular}{
m{1.2cm}                  
>{\centering\arraybackslash}m{1.7cm} 
>{\centering\arraybackslash}m{0.55cm} 
>{\centering\arraybackslash}m{0.55cm} 
>{\centering\arraybackslash}m{0.55cm} 
>{\centering\arraybackslash}m{0.65cm} 
>{\centering\arraybackslash}m{0.9cm} 
>{\centering\arraybackslash}m{0.6cm} 
>{\centering\arraybackslash}m{1.1cm} 
>{\centering\arraybackslash}m{0.8cm} 
>{\centering\arraybackslash}m{0.7cm} 
>{\centering\arraybackslash}m{0.65cm} 
>{\centering\arraybackslash}m{0.8cm} 
>{\centering\arraybackslash}m{0.9cm} 
>{\centering\arraybackslash}m{0.6cm} 
>{\centering\arraybackslash}m{1.2cm} 
>{\centering\arraybackslash}m{0.6cm} 
}
\toprule

& & \multicolumn{4}{c}{\textbf{Geometry Fidelity}}
& \multicolumn{4}{c}{\textbf{Dimension Fidelity}}
& \multicolumn{5}{c}{\textbf{Inference Efficiency}} 
& \multicolumn{2}{c}{\textbf{Iter. Stat.}} 
\\
\cmidrule(lr){3-6}
\cmidrule(lr){7-10}
\cmidrule(lr){11-15}
\cmidrule(lr){16-17}

\multirow{2}{*}{\textbf{Model}} &
\multirow{2}{*}{\textbf{Feedback}} &
\multirow{2}{*}{\makecell{\textbf{CD}}} &
\multirow{2}{*}{\makecell{\textbf{SIoU}}} &
\multirow{2}{*}{\makecell{\textbf{VIoU}}} &
\multirow{2}{*}{\makecell{\textbf{Surf.}\\\textbf{F-Scr.}}} &
\multirow{2}{*}{\makecell{\textbf{Surf.}\\\textbf{Err.}}} &
\multirow{2}{*}{\makecell{\textbf{Surf.}\\\textbf{Err.}}} &
\multirow{2}{*}{\makecell{\textbf{Vol.}\\\textbf{Err.}}} &
\multirow{2}{*}{\makecell{\textbf{Vol.}\\\textbf{Err.}}} &
\multirow{2}{*}{\makecell{\textbf{Infer.}\\\textbf{Time}}} &
\multirow{2}{*}{\makecell{\textbf{Pass}\\\textbf{Rate}}} &
\multirow{2}{*}{\makecell{\textbf{Input}\\\textbf{Tks.}}} &
\multirow{2}{*}{\makecell{\textbf{Output}\\\textbf{Tks.}}} &
\multirow{2}{*}{\makecell{\textbf{CAD}\\\textbf{Ops}}} &
\multirow{2}{*}{\makecell{\textbf{Self-Stop}\\\textbf{Rate}}} &
\multirow{2}{*}{\makecell{\textbf{Avg.}\\\textbf{Iter.}}} \\

\\
 & & $\downarrow$ & $\uparrow$ & $\uparrow$ & $\uparrow$ &
(mm$^2$)$\downarrow$ & (\%)$\downarrow$ &
(mm$^3$)$\downarrow$ & (\%)$\downarrow$ &
(s)$\downarrow$ & (\%)$\uparrow$ &
--- & --- & --- & (\%)$\uparrow$ & $\downarrow$ \\
\midrule
\multicolumn{17}{c}{\textbf{Commercial LLMs}} \\
\midrule

GPT& Adapt. Rndr. & 3.44 & 0.60 & 0.81 & 0.75 & 24298 & 9.0 & 616315 & 37.68 & 288.0 & 94.40 & 15879 & 15842 & 9.8 & \underline{99.80} & 2.75 \\
5.5 & Fixed 8 Rndr. & 2.92 & 0.63 & 0.82 & 0.77 & 22496 & 8.4 & 203721 & \textbf{15.91} & 217.6 & 92.37 & 29821 & 12850 & 9.8 & \underline{99.80} & 2.31 \\
\midrule

GPT& Adapt. Rndr. & 6.60 & 0.39 & 0.61 & 0.55 & 49616 & 20.7 & 2869741 & 131.81 & 48.5 & 80.77 & 43936 & 8203 & 9.8 & 43.00 & 7.73 \\
5.4 Mini & Fixed 8 Rndr. & 7.63 & 0.40 & 0.62 & 0.56 & 64983 & 21.3 & 2479906 & 95.74 & 76.1 & 77.07 & 115308 & 7998 & 9.9 & 40.40 & 7.83 \\
\midrule

Claude& Adapt. Rndr. & 4.18 & 0.53 & 0.74 & 0.69 & 35581 & 10.0 & 2435589 & 32.76 & 93.5 & \underline{95.09} & 38343 & 8420 & 10.1 & 97.80 & 3.54 \\
Opus 4.8 & Fixed 8 Rndr. & 4.14 & 0.51 & 0.73 & 0.67 & 24238 & 12.1 & 253217 & 31.20 & 99.0 & \textbf{96.93} & 40023 & 9550 & 9.9 & 99.20 & 4.17 \\
\midrule

Gemini& Adapt. Rndr. & 3.06 & 0.63 & 0.82 & 0.77 & 29560 & \textbf{7.4} & 1605669 & 35.64 & 381.2 & 82.31 & 11044 & 49609 & 13.5 & 99.60 & \underline{2.11} \\
3.1 Pro & Fixed 8 Rndr. & 3.54 & 0.62 & 0.82 & 0.76 & 30162 & 9.0 & 1469850 & 30.34 & 339.8 & 81.7 & 14771 & 45364 & 13.2 & 98.00 & \textbf{1.94} \\
\midrule

Gemini& Adapt. Rndr. & \textbf{2.70} & \textbf{0.63} & \textbf{0.83} & \textbf{0.77} & 21740 & 8.2 & 3480815 & 39.85 & 149.0 & 71.71 & 12559 & 36338 & 14.3 & \underline{99.80} & 2.50 \\
3.5 Flash & Fixed 8 Rndr. & \underline{2.81} & 0.62 & \underline{0.83} & 0.77 & 29483 & 7.5 & 2398220 & \underline{30.30} & 141.2 & 70.71 & 15343 & 32689 & 13.9 & \textbf{100.0} & 2.19 \\
\midrule

Qwen& Adapt. Rndr. & 4.77 & 0.50 & 0.71 & 0.66 & 35485 & 19.6 & 1786931 & 58.22 & 907.5 & 81.93 & 48059 & 48527 & 13.0 & 73.80 & 6.39 \\
3.7 Plus & Fixed 8 Rndr. & 4.69 & 0.47 & 0.70 & 0.64 & 37806 & 17.4 & 241526 & 54.21 & 934.6 & 79.90 & 71658 & 49112 & 13.3 & 83.80 & 5.41 \\
\midrule

\multicolumn{17}{c}{\textbf{Open-Source LLMs}} \\
\midrule

Qwen& Adapt. Rndr. & 3.25 & \underline{0.68} & 0.82 & \underline{0.81} & 13890 & \underline{7.4} & 683288 & 138.16 & 1270 & 5.18 & 42677 & 57711 & 9.1 & 1.40 & 4.02 \\
3.5 9b & Fixed 8 Rndr. & 8.74 & 0.38 & 0.59 & 0.54 & 278081 & 20.0 & 6239513 & 269.39 & 1729 & 4.67 & 56437 & 70450 & 12.1 & 3.00 & 4.80 \\
\midrule

Llama4& Adapt. Rndr. & -- & -- & -- & -- & -- & -- & -- & -- & 49.5 & 0.00 & 74407 & 6065 & 8.1 & 0.00 & 10.00 \\
Maverick & Fixed 8 Rndr. & -- & -- & -- & -- & -- & -- & -- & -- & 49.0 & 0.00 & 73697 & 5837 & 8.1 & 0.00 & 10.00 \\
\midrule

Ministral3& Adapt. Rndr. & -- & -- & -- & -- & -- & -- & -- & -- & 100.9 & 0.02 & 76805 & 13966 & 14.9 & 0.00 & 9.85 \\
4B 2512 & Fixed 8 Rndr. & 3.76 & 0.06 & 0.22 & 0.10 & 626 & 8.1 & 21505 & 177.80 & 107.0 & 0.02 & 76440 & 13952 & 15.5 & 0.00 & 9.98 \\

\midrule
\bottomrule


\end{tabular}
\caption{
Agentic Experiment Results (Sample N=2000). 
}
\label{tab:agentic_result}

\endgroup
\end{table*}

\subsection{Main Results and Findings}
The main results are summarized below, with more detailed results provided in Appendix~D.

\textbf{Real-World Requirements.}
In practical engineering workflows, visual resemblance alone is insufficient. 
The reconstructed CAD model must preserve specified features and dimensions; satisfy the tolerances required for downstream manufacturing and assembly. 
Consequently, even moderate geometric deviations or missing features may render a reconstruction unusable in practice.

\textbf{Pass Rate Criterion.}
A trial is successful only if the model returns a complete response and the extracted program produces a valid, non-empty 3D model file. 
Thus, the pass rate captures both response completeness and execution success. 
Missing responses and invalid or erroneous code are counted as failures; upon which evaluation metrics are not computed.

\textbf{Evaluation Results of Zero-shot and Agentic Study.} Tables~\ref{tab:oneshot_result} and~\ref{tab:agentic_result} summarize the zero-shot and agentic results respectively. Overall, both general-purpose VLMs and specialized models perform poorly on 3D reconstruction from mechanical drawings. The evaluation reveals persistent difficulties in interpreting orthographic views and dimensions, recovering feature relationships, and generating geometrically accurate CAD models. These findings highlight the limited capability of current models to comprehensively understand mechanical drawings and recover the exact design.

\textbf{Geometric Conformity.}
We evaluate shape agreement using Chamfer Distance (CD), Surface Intersection over Union (SIoU), Voxel Intersection over Union (VIoU), and Surface F-score. SIoU ranged from $0.042$ to $0.661$ and VIoU from $0.052$ to $0.844$, indicating substantial variation and limited conformity between generated and reference geometries. In real-world engineering, such deviations represent missing features and incorrect feature characteristic, rendering reconstructed parts unsuitable for assembly or manufacturing.

\textbf{Dimensional Fidelity.}
We evaluate dimensional accuracy using errors in surface area and volume. Surface-area error ranged from $7.09$--$27.26\%$, while volume error ranged from $17.41$--$145.82\%$, considering only commercial LLMs and across all ablations. In practical CAD workflows, dimensions must conform to specified values and tolerances; even visually minor errors can affect fit, function, and manufacturability. The observed errors therefore demonstrate a substantial gap between current reconstruction performance and real-world engineering requirements.

\textbf{Effect of Richer Visual Information.}
Additional views generally improved fidelity, but gains were model- and metric-dependent. Twenty renders often favored overlap metrics, whereas eight renders sometimes yielded better CD and dimensional accuracy. GPT-5.4 Mini showed the largest improvement: CD decreased from $7.25$ to $6.56$ and SIoU, VIoU, and F-score increased to $0.405$, $0.626$, and $0.569$, respectively, although pass rate dropped from $60.02\%$ to $48.90\%$. Video also improved Gemini~3.1 Pro’s overlap metrics, indicating that richer visual context helps non-monotonically and may reduce execution reliability.

\textbf{Agentic Versus Zero-shot Generation.}
Compared with the corresponding zero-shot settings, agentic generation primarily improved reliability rather than geometric accuracy. Qwen~3.7 Plus improved across all conformity metrics, while adaptive feedback raised GPT-5.4 mini’s pass rate from $55.48\%$ to $80.77\%$. Stronger models showed limited or negative geometric changes, suggesting that iterative feedback mainly improves execution success.



\textbf{Agentic Behavior vs. Model Scale.}
Frontier commercial models achieved self-stop rates of $97.8$--$100\%$ and substantially higher pass rates. In contrast, Qwen3.5-9B, Llama~4 Maverick, and Ministral-14B-2512 remained below $5.18\%$ pass rate and $3\%$ self-stop rate, with Llama~4 and Ministral often reaching the iteration limit without valid outputs. This indicates that agentic performance depends strongly on geometric reasoning, code reliability, and stopping calibration.

\textbf{Token Savings from Compressed Code History}
Diff compression reduced estimated code-history token usage by $42.44\%$, lowering multi-turn CAD refinement overhead while preserving prior implementation states. This benefit grows with session length, as repeatedly including complete scripts causes prompt size to scale poorly across iterations.

\textbf{Low Agentic Pass Rates of Smaller Models.}
Geometric metrics are omitted where no valid outputs were produced, as in both Llama~4 Maverick settings and adaptive Ministral 3 14B. These failures, caused primarily by incomplete or non-executable code, demonstrate that iterative visual feedback cannot compensate for insufficient model capability.

\textbf{Generation-length Limitation of CAD-Coder.}
In the author-released setting, CAD-Coder had an effective generation limit of approximately 4,096 tokens. Programs for complex parts often exceeded this limit and were truncated, reducing the pass rate as construction complexity increased. This limitation demonstrates that output capacity is a practical constraint in programmatic CAD reconstruction.

\textbf{Limitations of Image-to-3D Baselines.}
Hunyuan3D-2, InstantMesh, SAM~3D, and TRELLIS performed poorly on dimensioned engineering drawings, often treating the sheet as a single image and producing coarse, textured, or shallowly extruded geometry rather than the specified part. This reflects the gap between appearance-oriented reconstruction and engineering CAD reconstruction, which requires reasoning across orthographic views, dimensions, hidden geometry, constraints, and parametric dependencies.

\section{Conclusion}
We introduce \textit{\name}, a benchmark for evaluating LMMs on 2D engineering drawing to 3D CAD reconstruction, consisting of 251K real-world mechanical drawing tasks paired with reference 3D models and structured geometric evaluation protocols.
We evaluate state-of-the-art LMMs under image-only, render-augmented, video-augmented, and agentic settings, alongside representative 3D reconstruction baselines.
Our analysis reveals critical limitations of LMMs, including weak dimension grounding, incomplete integration of orthographic and sectional views, poor reconstruction of features, and generated CAD models that deviate from reference designs across global feature, dimensional accuracy, surface geometry, volume, area, and mesh-level similarity.

\bibliography{aaai2027}

\end{document}